\documentclass{article}
\pdfoutput=1
    \PassOptionsToPackage{numbers, compress}{natbib}

    \usepackage[final]{neurips_2020}
    \usepackage{neurips_2020}

\usepackage[utf8]{inputenc} 
\usepackage[T1]{fontenc}    
\usepackage{hyperref}       
\usepackage{url}            
\usepackage{booktabs}       
\usepackage{amsfonts}       
\usepackage{nicefrac}       
\usepackage{microtype}      
\usepackage[ruled,vlined]{algorithm2e}
\usepackage{xcolor}
\usepackage{amsmath}
\DeclareMathOperator*{\argmax}{arg\,max}

\usepackage{textcomp,gensymb}
\usepackage{multirow}
\usepackage{graphicx}
\usepackage{siunitx}
\usepackage{amsmath,bm}
\newlength\myindent
\author{
Vidhi Jain\\
Carnegie Mellon University \\
\texttt{vidhij@andrew.cmu.edu}

\And
Prakhar Agarwal* \\
University of Washington, Seattle \\
\texttt{pa2511@u.washington.edu}

\And
Shishir Patil*  \\
University of California, Berkeley\\
\texttt{shishirpatil@berkeley.edu}

\And
Katia Sycara \\
Carnegie Mellon University \\
\texttt{katia@cs.cmu.edu}
}

\title{Learning Embeddings that Capture Spatial Semantics for Indoor Navigation}

\begin{document}

\maketitle

\begin{abstract}

Incorporating domain-specific priors in search and navigation tasks has shown promising results in improving generalization and sample complexity over end-to-end trained policies. 
In this work, we study how object embeddings that capture spatial semantic priors can guide search and navigation task in a structured environment. 
We know that humans can search for an object like a book, or a plate in an unseen house, based on spatial semantics of bigger objects detected. For example, a book is likely to be on a bookshelf or a table, whereas a plate is likely to be in a cupboard or dishwasher. We propose a method to incorporate such spatial semantic awareness in robots by leveraging pre-trained language models and multi-relational knowledge bases as object embeddings. We demonstrate using these object embeddings to search a query object in an unseen indoor environment. We measure the performance of these embeddings in an indoor simulator (AI2Thor).
We further evaluate different pre-trained embedding on \textit{Success Rate} (SR) and \textit{Success weighted by Path
Length} (SPL). \\ 
Code is available at: \href{https://github.com/vidhiJain/SpatialEmbeddings}{\color{blue}{https://github.com/vidhiJain/SpatialEmbeddings}}


\end{abstract}

\section{Introduction}




Consider an example of finding a key-chain in a living room. A key-chain is found either on a coffee table, inside a drawer, or on a side-table. When tasked with finding the key-chain, a human would first scan the area coarsely and then navigate to likely locations where a key-chain could be found, for example, a coffee table. On getting closer, they would then examine the area (top of the table) closely. Following this, if the key-chain is not found, they would try to navigate to the next closest place where the key-chain could be found. In all these scenarios, the presence (or absence) of a co-located objects would boost (or dampen) their confidence in finding the object along the chosen trajectory. This would, in turn, influence the next step (action) that they would take. 

Our goal is to enable embodied AI agents to navigate based on such object-based spatial semantic awareness. To do this, we focus on the following problems: (a) training object embeddings that semantically represent spatial proximity, and (b) evaluating these embeddings on semantic search and navigation tasks. We aim to learn embeddings that capture the following - (a) learn about larger objects around which the smaller items could be found (e.g., key-chains are likely to be found on / near tables), and (b) learn about objects that are found mutually close to each other, e.g.,  (key-chains and credit-card). By learning such embeddings, we want to capture the semantic relations in terms of distance. 
We train embeddings that capture the spatial semantics using a multi-relational knowledge graph. 
Further, we formulate an algorithm to compare the performance 
in terms of \textit{Success Rate} (SR) and \textit{Success weighted by Path Length} (SPL) 
across different kinds of embeddings. 
Interestingly, embeddings trained with general-purpose text corpora like Word2Vec \cite{word2vec}, FastText \cite{fasttext1} provide a strong baseline for our task. 
We demonstrate our techniques on the AI2Thor \cite{robothor} environment. 




Our contributions are summarized as follows: (a) We learn and contrast representations to capture spatial semantics.
(b) We demonstrate that a momentum inspired, similarity-based greedy navigation technique results in success rate $>90\%$. (c) Finally, we  show that using spatial semantic prior knowledge can significantly improve the navigation performance.


\begin{figure}
    \centering
    \includegraphics[width=\columnwidth]{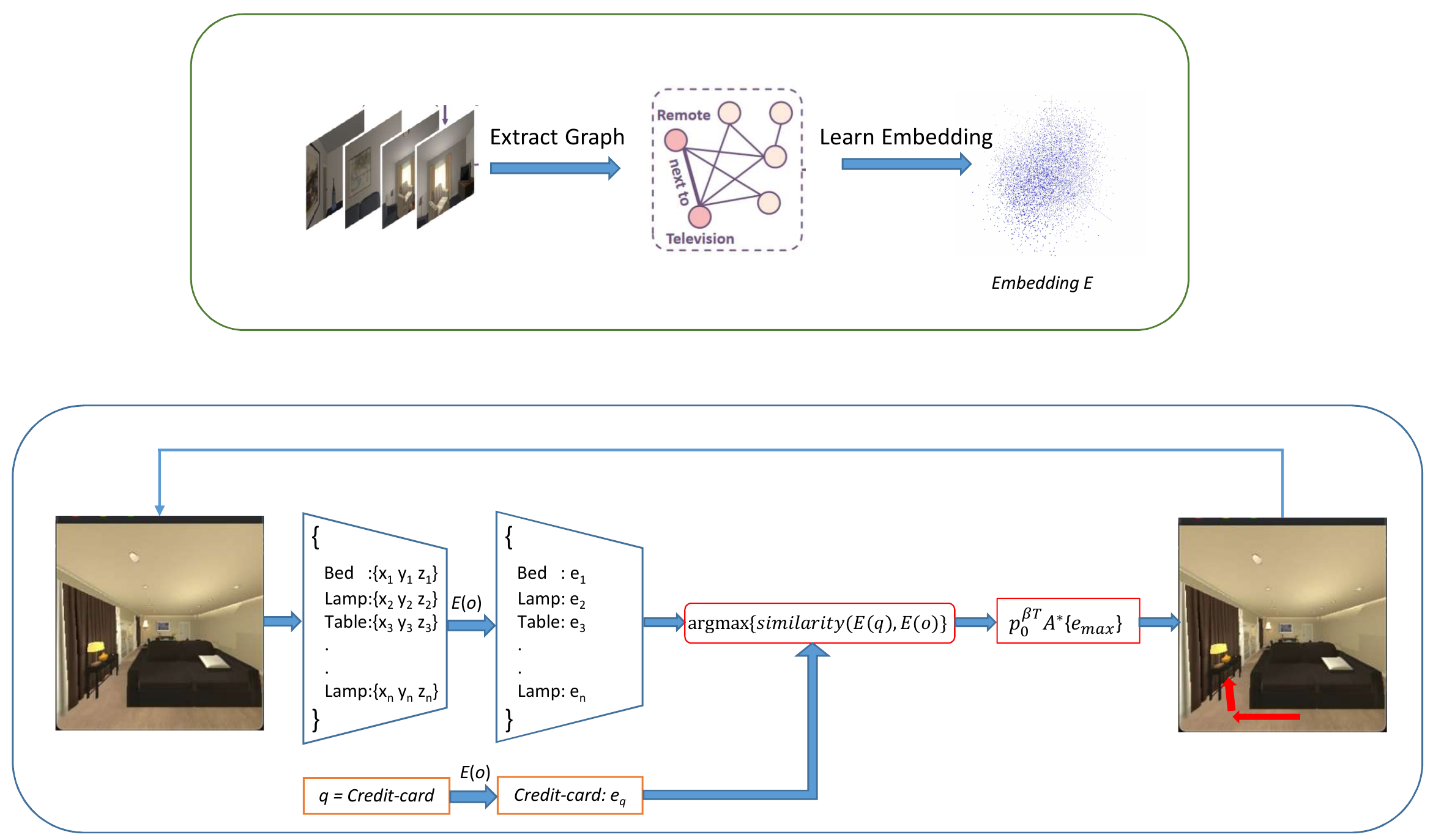}
    \caption{[Top] Training the embeddings, [Bottom] Using embeddings for navigation}
    \label{fig:architecture}
\end{figure}


\section{Related Work}

Classical navigation techniques in robotics generalize to unseen environments by mapping, localization, and path planning. However, most of these approaches fail to leverage the environment's semantic structure, like certain objects are mutually close to each other.

Multi-relational knowledge-base based embeddings such as RoboCSE \cite{daruna2019robocse} have demonstrated an agent's ability to predict object affordances and materials in the real world. The RoboCSE embeddings were trained using the object metadata from AI2Thor simulator. While these embeddings were used for the prediction of object affordances by an immovable agent, we differ in our approach of utilizing these embeddings by evaluating how well they capture the spatial semantics for query search and navigation task for a mobile agent.





Prior work has also looked at using object embeddings for robot navigation.
Graph convolutional networks have been used to learn neural-network based policy in
\cite{niko2019keys} and  \cite{DBLP:journals/corr/abs-1810-06543} for object search and navigation. As these are data-driven approaches, they require several interactions in the environment to learn the policy. We propose a sample-efficient approach based on pre-trained embedding [Sec ~\ref{sec:methodology}]. While one approach is to learn an end-to-end navigation policy from scratch, we demonstrate the promise of using embeddings for navigation, and their transfer to real-world settings.

Recent work in object navigation \cite{chaplot2020object} shows a modular approach to map the environment and predict sub-goal based on the semantic object map. We do not address the vision-based mapping as addressed in this paper. Instead, our approach is a modular component that can be integrated with classic robotics paradigm of mapping and planning. We focus on high-level navigation decisions aligned to the spatial semantics of objects. 







\begin{algorithm}[t]
\SetAlgoLined
\KwResult{Plan to the query object}
 embedding module E\;
 visited = []\;
 \While{ query $q$ not found}{
  \For{ all objects $o$ in (360 degree) field of view}{
      \If{object o not in visited}{
       $\text{scores} = \text{similarity}(E(q), E(o))$\;
       Add $o$ to potential\_sub\_goals\;
      }
  }
  sub-goal $= \text{potential\_sub\_goals}[\argmax \text{scores}$]\;
  proposed plan  $p_0^T := A^*$ search to sub-goal\;
  move according to initial part of the proposed plan $p_0^{\beta T}$ where $\beta < 1$\;
 }
 \caption{Navigation towards query for the objects in the field of view}
 \label{algorithm1}
\end{algorithm}
\section{Task Definition}
Our goal is to navigate an agent from a randomly initialized location in a scene to a specified query object. The task is a success if the agent can view the object of interest, and is within a small distance ($\sim$ 1 units $= 5 \times 0.25$) of the query object. 

Consider the agent is tasked to find a \textit{credit-card} (query). The agent is initialized randomly in a living room and records all the objects in its 360\degree~ field of view. By design, we enforce that the agent can only see ``large'' objects.
Imagine you enter a living room and see a sofa, table, TV, etc. easily. 
However, depending on your distance, objects such as 
the credit card might be occluded or not be visible. We consider large objects that occupy more than 1\% of the total screen size of the camera's image. This threshold is analogous to a real camera's effects as detection of objects in the frame would depend on the camera's resolution.



The agent now finds a sub-goal. The sub-goal is the most likely (large) object in its field of view, which we expect is close to the query object,  {or} could lead the agent towards a viewpoint that brings relevant sub-goals in the agent's field of view. 
This is done by finding the pairwise similarity of the query object's embedding with the embeddings of all the sufficiently large objects in sight. 
For example, out of visible objects like table, lamp, and bed, we would expect embeddings of a table to have the highest similarity with the embeddings of a credit-card. 

Once we choose a sub-goal, we calculate the potential plan $p_0^{T_i}$ where $T_i$ is the total steps required to reach the sub-goal $i$. The agent executes
$p_0^{\beta T}$ that is some fraction of the initial part of the potential plan. The agent then looks for any new object that is now visible and re-ranks the similarity of all objects visible to it to find the new sub-goal. This process repeats until the query object is found.
Consider that in a 1D space, the agent's location is at 0, and the query object's location is at 10. With $\beta=0.5$, the agent would first navigate to point 5, then to 7, and finally to 9 before reaching 10. 
This approach has two advantages: a) improves the performance of the system by exponentially reducing the amount of compute per navigation (scan, similarity computation, $A^*$ path computation). b) helps overcome oscillation. Oscillations in navigation happen when the agent takes a step towards sub-goal A, and finds B to be having a higher similarity. On shifting the sub-goal, and taking a step towards B, B becomes occluded, and the sub-goal rotates back to A (Appendix~\ref{navosc})    


\section{Methodology}
\label{sec:methodology}


Our novel approach of finding the semantic similarity between the given query and visible objects is formulated based on the distribution of distance between a pair of objects. Additionally, we identified two broad kinds of embeddings for our analysis, as discussed below. Each of these embeddings are evaluated for navigation as outlined in Algorithm \ref{algorithm1}. 

\textbf{Pre-trained word embeddings}
Language embeddings capture the word-level semantics in their metric space. For example, FastText embeddings have shown promising results for semantic navigation in procedurally generated environments \cite{niko2019keys}. To understand if these embeddings can be transferred in terms of object semantics 
for the downstream task of navigation
, we test our algorithm on the Word2Vec \cite{word2vec} and FastText \cite{fasttext1} embeddings. 

\textbf{Knowledge base embeddings}
Multi-relational embeddings encode abstract knowledge that could be obtained by the agent from its sensors or an external knowledge graph. RoboCSE \cite{daruna2019robocse} uses ANALOGY\cite{liu2017analogical}, a semantic matching
method, to learn multi-relational embeddings processed from the AI2Thor environment data, where the nodes represent the objects and edges denote the relation between them.
Further, we also learn a Graph Embedding by treating all relations as being equivalent and using DeepWalk \cite{perozzi2014deepwalk} to learn embedding for the resulting undirected graph.

\section{Experiment and Results}
\label{experiment}

We use the AI2Thor environment to extract pairwise object relations 
to train the embedding network. We make two interesting design choices that we think will be valuable to the community. First, since we assume perfect object detection if the object is sufficiently in the field of view, we determine sufficiency as a parameter with respect to the percentage of pixels occupied by the object in the field of view. 
Second, we adaptively choose step-sizes to navigate towards the sub-goal. This allows the agent efficiently to look for other objects in the field of view on the way. For more information refer Appendix~\ref{Appendix}.

In Table \ref{tab:results}, we show the performance of different embeddings for navigation task in terms of Success weighted by normalized inverse path length (SPL) metric. Further, we also report SPL computed by grouping cases where the shortest path length to the query object is within a particular range. For example, the credit card may be initialized (randomly) 15m away from the agent, including it in the $10<l<20$ bucket. 

We observe that the \textbf{graph-based embeddings (RoboCSE and Graph Embeddings) outperform the baselines in cases where the optimal path to the target object is higher} ($10<l<20, l>=20$) but perform comparably to language-based embeddings (Word2Vec, FastText) when optimal paths are shorter ($l<10$). Our hypothesis is that, for a query object close  to the agent's initialisation ($l<10$), any action that the agent takes has a higher probability of taking it closer to the query object. However, for objects that are further away, the agent benefits from determining the most likely path to get to the query object.

\begin{table}[]
\begin{tabular}{clccrr}
\hline
\multicolumn{1}{l}{} &
  \multicolumn{1}{c}{} &
  \multicolumn{1}{l}{} &
  \multicolumn{3}{c}{SPL by shortest path length $l$} \\ \cline{4-6} 
\multicolumn{1}{l}{Room Type} &
  \multicolumn{1}{c}{Embedding Type} &
  SR/SPL &
  $l$< 10 &
  \multicolumn{1}{c}{10 <$l$< 20} &
  \multicolumn{1}{c}{$l$ >= 20} \\ \hline
\multicolumn{1}{c|}{\multirow{4}{*}{Kitchen}} &
  \multicolumn{1}{l|}{Graph Embedding} &
  \multicolumn{1}{c|}{\textbf{{0.992} \hfill / \hfill {0.639}}} &
  {0.644} &
  {0.642} &
  {0.644} \\
\multicolumn{1}{c|}{} &
  \multicolumn{1}{l|}{RoboCSE} &
  \multicolumn{1}{c|}{{0.960}\hfill / \hfill{0.624}} &
  \textbf{{0.650}} &
  {0.643} &
  \textbf{{0.867}} \\
\multicolumn{1}{c|}{} &
  \multicolumn{1}{l|}{FastText} &
  \multicolumn{1}{c|}{{0.983}\hfill / \hfill{0.615}} &
  {0.626} &
  {0.624} &
  {0.573} \\
\multicolumn{1}{c|}{} &
  \multicolumn{1}{l|}{Word2Vec} &
  \multicolumn{1}{c|}{{0.984}\hfill / \hfill{0.626}} &
  {0.633} &
  \textbf{{0.649}} &
  {0.581} \\ \hline
\multicolumn{1}{c|}{\multirow{4}{*}{Living Room}} &
  \multicolumn{1}{l|}{Graph Embedding} &
  \multicolumn{1}{c|}{{0.919}\hfill / \hfill{0.692}} &
  {0.777} &
  {0.698} &
  \textbf{{0.693}} \\ 
\multicolumn{1}{c|}{} &
  \multicolumn{1}{l|}{RoboCSE} &
  \multicolumn{1}{c|}{\textbf{{0.942}\hfill / \hfill{0.686}}} &
  {0.766} &
  {0.692} &
  {0.614} \\
\multicolumn{1}{c|}{} &
  \multicolumn{1}{l|}{FastText} &
  \multicolumn{1}{c|}{{0.908}\hfill / \hfill{0.682}} &
  {0.774} &
  \textbf{0.727} &
  {0.619} \\
\multicolumn{1}{c|}{} &
  \multicolumn{1}{l|}{Word2Vec} &
  \multicolumn{1}{c|}{{0.908}\hfill / \hfill{0.666}} &
  \textbf{{0.793}} &
 {0.708} &
  {0.596} \\ \hline
\multicolumn{1}{c|}{\multirow{4}{*}{Bed Room}} &
  \multicolumn{1}{l|}{Graph Embedding} &
  \multicolumn{1}{c|}{{0.954}\hfill / \hfill{0.678}} &
  \textbf{{0.739}} &
  {0.631} &
  {0.659} \\
\multicolumn{1}{c|}{} &
  \multicolumn{1}{l|}{RoboCSE} &
  \multicolumn{1}{c|}{\textbf{{0.966}\hfill / \hfill{0.681}}} &
  {0.731} &
  {0.628} &
  \textbf{{0.690}} \\
\multicolumn{1}{c|}{} &
  \multicolumn{1}{l|}{FastText} &
  \multicolumn{1}{c|}{{0.960}\hfill / \hfill{0.686}} &
  {0.738} &
  \textbf{{0.657}} &
  {0.544} \\
\multicolumn{1}{c|}{} &
  \multicolumn{1}{l|}{Word2Vec} &
  \multicolumn{1}{c|}{{0.956}\hfill / \hfill{0.662}} &
  {0.720} &
  {0.624} &
  {0.576} \\ \hline
\multicolumn{1}{c|}{\multirow{4}{*}{Bath Room}} &
  \multicolumn{1}{l|}{Graph Embedding} &
  \multicolumn{1}{c|}{{0.997}\hfill / \hfill{0.694}} &
  {0.692} &
  {0.733} &
  \multirow{4}{*}{-} \\
\multicolumn{1}{c|}{} &
  \multicolumn{1}{l|}{RoboCSE} &
  \multicolumn{1}{c|}{{0.994}\hfill / \hfill{0.692}} &
  {0.693} &
  {0.716} &
   \\
\multicolumn{1}{c|}{} &
  \multicolumn{1}{l|}{FastText} &
  \multicolumn{1}{c|}{{0.994}\hfill / \hfill{0.688}} &
  {0.692} &
  {0.690} &
   \\
\multicolumn{1}{c|}{} &
  \multicolumn{1}{l|}{Word2Vec} &
  \multicolumn{1}{c|}{\textbf{{0.998}\hfill / \hfill{0.701}}} &
  \textbf{{0.697}} &
  \textbf{{0.743}} &
  \\ \hline 
\end{tabular}
\caption{\label{tab:results}\textbf{Results for navigation using Algorithm~\ref{algorithm1}.} Success Rate (SR) and Success weighted normalized inverse path length (SPL) for different floor plans in AI2thor environment; $l$ denotes the shortest path length to query object.}
\end{table}

\newpage

\section*{Future Work}
\label{experiment}

In our work, when the agent is tasked with navigating to $n, (n>1)$ objects, there are two possible scenarios, a) all $n$ objects are given as queries at the start, b) objects are queried sequentially one-by-one. For addressing (a), Algorithm ~\ref{algorithm1} can be modified to jointly optimize for all the $n$ objects. For addressing (b), it involves handling consecutive searches based on the information gained during navigation for the first $(n-1)$ objects to better navigate the $n^{th}$ object. 
Both of these are avenues for future work.
\section*{Broader Impact}
Our approach is to learn an embedding that encapsulates the relationship between the smaller query object (like an apple) and their possible parent receptacles (like a refrigerator). We believe this is a first step towards robust navigation for embodied AI agents. We further believe that a extensive system built with the principles defined here would enable efficient techniques for navigation tasks. 

\begin{ack}
This material is based upon work supported by the Defense Advanced Research Projects Agency (DARPA) under Contract No. HR001120C0036. Any opinions, findings and conclusions or recommendations expressed in this material are those of the author(s) and do not necessarily reflect the views of the Defense Advanced Research Projects Agency (DARPA).
\end{ack}


\bibliographystyle{unsrt}
\bibliography{sample}

\begin{thebibliography}{1}

\bibitem{word2vec}
Tomas Mikolov, Kai Chen, Greg Corrado, and Jeffrey Dean.
\newblock Efficient estimation of word representations in vector space, 2013.

\bibitem{fasttext1}
Piotr Bojanowski, Edouard Grave, Armand Joulin, and Tomas Mikolov.
\newblock Enriching word vectors with subword information.
\newblock {\em arXiv preprint arXiv:1607.04606}, 2016.

\bibitem{robothor}
Matt Deitke, Winson Han, Alvaro Herrasti, Aniruddha Kembhavi, Eric Kolve,
  Roozbeh Mottaghi, Jordi Salvador, Dustin Schwenk, Eli VanderBilt, Matthew
  Wallingford, Luca Weihs, Mark Yatskar, and Ali Farhadi.
\newblock {RoboTHOR: An Open Simulation-to-Real Embodied AI Platform}.
\newblock 2020.

\bibitem{daruna2019robocse}
Angel Daruna, Weiyu Liu, Zsolt Kira, and Sonia Chernova.
\newblock Robocse: Robot common sense embedding, 2019.

\bibitem{niko2019keys}
Niko Sünderhauf.
\newblock Where are the keys? -- learning object-centric navigation policies on
  semantic maps with graph convolutional networks, 2019.

\bibitem{DBLP:journals/corr/abs-1810-06543}
Wei Yang, Xiaolong Wang, Ali Farhadi, Abhinav Gupta, and Roozbeh Mottaghi.
\newblock Visual semantic navigation using scene priors.
\newblock {\em CoRR}, abs/1810.06543, 2018.

\bibitem{chaplot2020object}
Devendra~Singh Chaplot, Dhiraj Gandhi, Abhinav Gupta, and Ruslan Salakhutdinov.
\newblock Object goal navigation using goal-oriented semantic exploration,
  2020.

\bibitem{liu2017analogical}
Hanxiao Liu, Yuexin Wu, and Yiming Yang.
\newblock Analogical inference for multi-relational embeddings.
\newblock {\em arXiv preprint arXiv:1705.02426}, 2017.

\bibitem{perozzi2014deepwalk}
Bryan Perozzi, Rami Al-Rfou, and Steven Skiena.
\newblock Deepwalk: Online learning of social representations.
\newblock In {\em Proceedings of the 20th ACM SIGKDD international conference
  on Knowledge discovery and data mining}, pages 701--710, 2014.

\end{thebibliography}

\pagebreak

\section*{Appendix}
\label{Appendix}

\subsection{Embedding Training}
\label{EmbTrain}
We describe our technique to train our Graph Embeddings. We used standard off the shelf pre-trained Word2Vec (trained on Google News) and FastText embeddings (trained on en Wikipedia), with no modifications. RoboCSE \cite{daruna2019robocse} uses ANALOGY\cite{liu2017analogical}, a semantic matching
method, to learn multi-relational embeddings processed from the AI2Thor environment data, where the nodes represent the objects and edges denote the relation between them.

To train the Graph Embeddings, we start by extracting a knowledge graph using a subset of floor plans for each Room type in the AI2Thor environment. The knowledge graph captures the co-location of objects and the relationship between them. Eg. \textit{Book} is \textbf{located} in the living room \textbf{on} a \textit{table}. The nodes in the graph represents objects (eg. Book, Table) and edges represents relationship (eg. on).
We then used Deepwalk (number of walks: 20, walk length: 8 and window size: 3) to generate a 25 dimensional embedding.
Once trained we do not modify the embeddings during navigation.

\subsection{Navigation: Oscillations}
\label{navosc}
Navigating to the object by choosing the object with the highest cosine similarity (with respect to \textit{query object}) could potentially suffer from oscillations. For example,consider the agent is at position A, and is navigating to position B. On taking a few steps towards position B, the object at position B could potentially become occluded. Then the object would navigate towards the (next visible) object with highest cosine similarity. This could be at position C. Now on taking a few steps towards position C, object at position B becomes visible again. This cycle repeats. In algorithm ~\ref{algorithm1}, the agent takes $p^{\beta T}_{0}$ steps at a time, which is some fraction of steps of the plan. While this heuristic does not alleviate the possibility of oscillations, in our experiments, we observed negligible oscillations. 

\subsection{Hyperparameters for Reproducibility}
\label{hyper}
The code is shared at \href{https://github.com/vidhiJain/SpatialEmbeddings}{\color{blue}{https://github.com/vidhiJain/SpatialEmbeddings}} 


Here, we report the hyperparameters which can be used to replicate the results presented in Table~\ref{tab:results}.

\texttt{VISIBLE AREA THRESHOLD = 0.01}; Objects that are lower than this ratio, are discarded. 

\texttt{STEP MULTIPLE = 5}; How many \texttt{``Steps * 0.25 units''} to move in each time step.

\texttt{Random Seed: 33} to initialize agent in AI2Thor and randomize the object initialization in the environment.

\end{document}